\pdfoutput=1
\documentclass[runningheads]{llncs}
\usepackage{graphicx}
\usepackage{amsmath,amssymb} 
\usepackage{color}
\usepackage[width=122mm,left=12mm,paperwidth=146mm,height=193mm,top=12mm,paperheight=217mm]{geometry}
\usepackage{times}
\usepackage{comment}
\usepackage{epsfig}
\usepackage{graphicx}
\usepackage{amsmath}
\usepackage{amssymb}
\usepackage{epstopdf}
\usepackage{subfig}
\usepackage[table]{xcolor}
\usepackage{xcolor,colortbl}
\usepackage{booktabs}
\usepackage{adjustbox}
\usepackage{array,multirow,graphicx}
\usepackage{cite}
\usepackage{multirow}
\usepackage{bigstrut}
\usepackage{authblk}
\usepackage[T1]{fontenc}
\usepackage[utf8]{inputenc}

\definecolor{TableBorder}{HTML}{303030}
\definecolor{TableBlack}{HTML}{000000}
\definecolor{TableRed}{HTML}{800000}
\definecolor{TableGreen}{HTML}{006633}
\definecolor{TableEven}{HTML}{D0D0D0 }
\definecolor{TableOdd}{HTML}{FFFFFF}
\definecolor{TableLightGray}{HTML}{E8E8E8}

\newcommand\VRule[1][\arrayrulewidth]{\vrule width #1}
\newcommand{\TextBorder}[1]{\textcolor{TableBlack}{\textbf{#1}}}%
\newcommand{\TextRed}[1]{\textcolor{TableRed}{#1}}
\newcommand{\TextGreen}[1]{\textcolor{TableGreen}{{#1}}}
\newcommand{\CellLightGray}{\cellcolor{TableLightGray}}

\mainmatter
\def\ECCV16SubNumber{864}  

\title{Friction from Reflectance: Deep Reflectance Codes for Predicting Physical Surface Properties from One-Shot In-Field Reflectance} 

\titlerunning{Deep Reflectance Codes}

\authorrunning{H. Zhang {\it et al.}}

\author{
Hang Zhang$^1$ Kristin Dana$^1$  Ko Nishino$^2$}
\institute{$^1$Department of Electrical and Computer Engineering, Rutgers University\\
$^2$Department of Computer Science, Drexel University\\
{\tt\small zhang.hang@rutgers.edu, kdana@ece.rutgers.edu,  kon@drexel.edu}
}

\begin{document}
\maketitle

\begin{abstract}
Images are the standard input for vision algorithms, but one-shot in-field reflectance measurements are creating new opportunities for recognition and scene understanding. 
In this work, we address the question of what reflectance can reveal about materials in an efficient manner.
We go beyond the question of recognition and labeling and ask the question:
What intrinsic physical properties of the surface can be estimated using reflectance? We introduce a framework that enables prediction of actual friction values for surfaces using one-shot reflectance measurements. This work is a first of its kind vision-based friction estimation. 
We develop a novel representation for reflectance disks that capture partial BRDF measurements instantaneously. Our method of {\it deep reflectance codes} combines CNN features and fisher vector pooling with optimal binary embedding to create codes that have sufficient discriminatory power and have important properties of illumination and spatial invariance. 
The experimental results demonstrate that reflectance can play a new role in deciphering the underlying physical properties of real-world scenes.

  
\end{abstract}

\section{Introduction}
Reflectance describes the characteristics of light interaction with a surface, which is uniquely determined by how the surface is made up at a microscopic (e.g., pigments in the surface medium) and mescopic scale (e.g., geometric 3D texture).  Naturally, reflectance provides an invaluable clue about the surface, including what it is made of (i.e., material), and how it is shaped (i.e., surface roughness). Reflectance has recently been used to classify surfaces based on the underlying materials \cite{Zhang15,Liu_2013_CVPR}.
and to recover fine-grained geometry \cite{Dana04,Wang06} 
Reflectance, however, encodes richer information about the surface than just the material category and surface geometry.

In this paper, we show that reflectance can be used to estimate tactile properties of the surface: determine the touch from appearance. We, in particular, build the first approach for {\it friction-from-reflectance}. An image-based friction estimation method has important implications in many applications. For example, robotics can plan how to grasp an object easily with an estimated friction coefficient. A mobile robot or autonomous automobile can use friction information to select an optimized driving mode. For this, we introduce a novel representation of reflectance patterns using a state-of-the-art texture representation that builds on key aspects of deep learning and  binary embedding. 
We demonstrate the power of our approach {\it deep reflectance codes} first in conventional material recognition and then on predicting friction from one-shot measurements.



\begin{figure}[t]
\centering
\begin{adjustbox}{max width=\textwidth}
\includegraphics[]{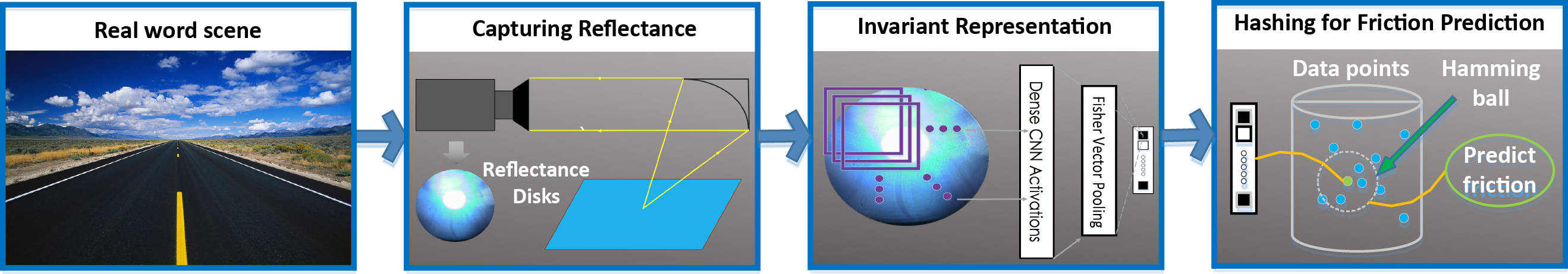}
\end{adjustbox}
\caption{ 
Deep reflectance codes for {\it friction-from-reflectance}. Reflectance is captured using one-shot in-field measurements. The binary embedding of Fisher Vector CNN preserves physical properties and provides invariant representation. The resulting hash codes is used in prediction of surface friction. 
}
\label{fig:diskexample}
\end{figure}

Reflectance measurements encode rich information that raw images do not. Recognizing materials using ordinary internet-mined photographs has shown great promise \cite{bell15minc,Sharan13,Cimpoi14}. Reflectance information provides a different, complementary approach \cite{Zhang15} based on the measurement of the material characteristics, instead of
 phenomenological appearance. As first introduced by Zhang et al.\cite{Zhang15}, reflectance disks are a sampling of a surface BRDF without requiring a lab based dome-like measurement system. The disk images are acquired 
using a concave parabolic mirror positioned over a surface. Due to the mirror geometry,  each pixel corresponds to the reflected surface light from a different viewing direction.

Computational models of reflectance disks need to be sufficiently descriptive and yet have invariance to illumination and surface tilt. The measured appearance of a surface changes under multiple illumination angles. 
The illumination direction affects the position of the specular peak within the reflectance disk, necessitating translation invariance within the representation. The distribution of patterns in the reflectance disk also necessitates a shift-invariant representation similar to desired properties of texture representations. 
Prior work in representing reflectance disks \cite{Zhang15} used hashing to create a binary code for fast recognition and retrieval in high dimensional spaces. 
Classic methods of texton histograms and texton maps have been replaced in recent years with deep representations for texture such as deep filter banks \cite{Cimpoi15}. We introduce a novel material representation, which we refer to as \textit{deep reflectance codes (DRC)}, that builds on these two concepts. Deep reflectance codes achieve necessary descriptive power and invariance by applying binary embedding to fisher vector convolutional neural nets (FV-CNN) \cite{Cimpoi15}. 
To demonstrate the general discriminatory power of this new material representation, we test it in both image-based and reflectance disk-based material recognition and show that their performance surpasses past representations on several existing material datasets. Figure~\ref{fig:diskexample} shows an overview of our approach.

\begin{figure}[!t]
\includegraphics[width=\linewidth]{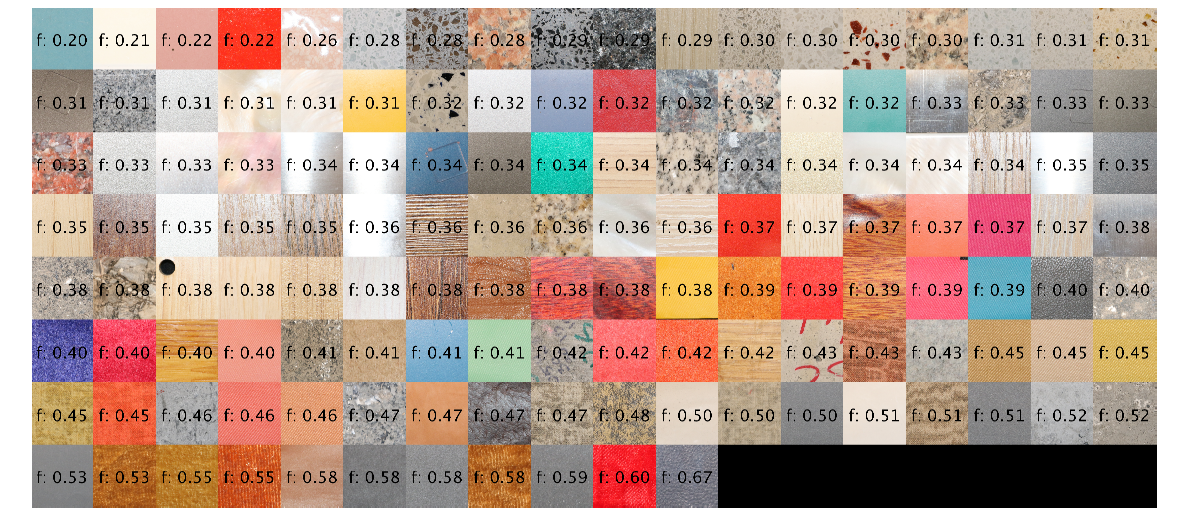}\\
\includegraphics[width=\linewidth]{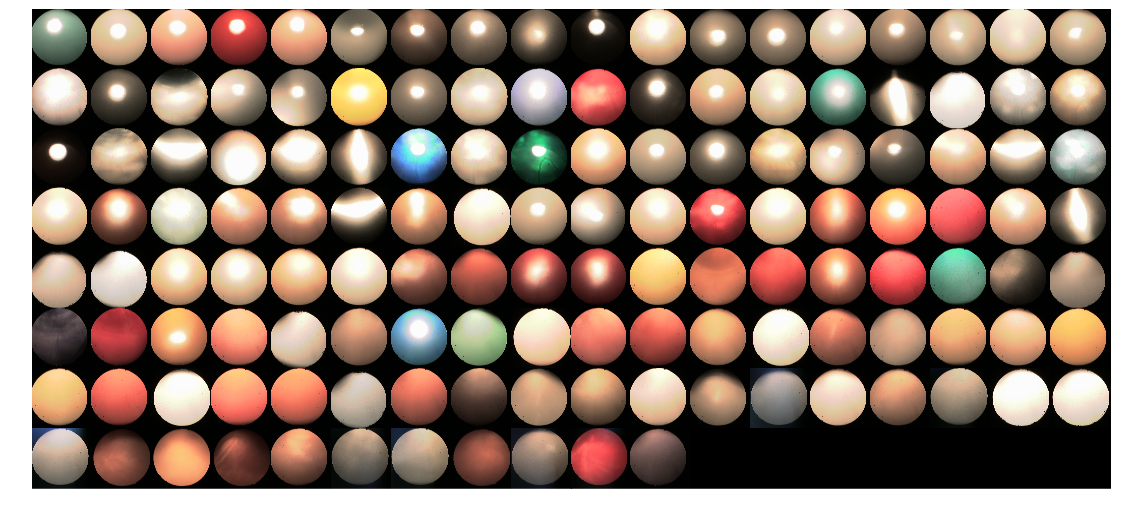}
\caption{
(Top) Images of the 137 material surfaces for the friction-material database ordered by friction coefficient. (Bottom) Example reflectance disks for corresponding material surfaces. The names for all material classes and the number of instances captured per class are shown in Table~\ref{fig:frictionsensor}.  As examples of  material class names, the list of names for the first surface in each row are as follows: Smooth Ceramic Tile, Automotive Paint, Marble, Composite Flooring, Asphalt, Nylon, Linen, Leather and Sand Paper. As expected, samples such as Smooth Ceramic Tile have lower coefficients of friction (0.2) than samples such as Sand Paper (0.53).}
\label{fig:database}
\end{figure}

Deep reflectance codes provide a compact binary representation of the intrinsic phyiscal characteristics of the underlying material. We demonstrate this by showing, for the first time, that reflectance can be used to estimate the friction coefficient of a surface. For this, we collect a first-of-its-kind database of friction coefficients and reflectance disks measured for 137 surfaces that can be grouped into 21 classes (shown in Figure~\ref{fig:database}). For exploratory analysis of how reflectance disks encode surface friction, we find a manifold showing the proximal layout of  deep reflectance codes using t-SNE \cite{van2008visualizing} as well as a corresponding friction map in t-SNE space. The experimental results show that the reflectance disks and their deep codes encode sufficient information to accurately predict the friction of a surface from its one-shot in-field reflectance measurements. The practical implications of this novel vision-based friction estimation is significant.
Current friction sensors are tactile such as tire sensors that measure slip \cite{Erdogan11,Gustafsson97,Matsuzakia05}. Appearance modeling in a fine-grained precise manner allows non-contact friction estimation (e.g. before the tire hits the road patch), which would be useful for various applications such as road surface ice detection, robot locomotion control, and the integration of haptics and graphics.

In summary, there are three main novel contributions: 1) deep reflectance codes for material representation, 2) friction estimation from reflectance, 3) and a database of friction coefficients and reflectance disks.

\section{Related Work}


Friction from reflectance is an entirely new area. Prior work in friction measurement requires surface contact.
These contact friction sensors have been used for real-world friction estimates in applications such as haptics for textiles \cite{ramkumar2003developing, bertaux2007relationship}, automobile  tire sensors \cite{Erdogan11,Gustafsson97,Matsuzakia05}, and sheet metal rolling in manufacturing \cite{nyahumwa1991friction}.
The ability to estimate friction with vision-based methods will have significant impact on these application areas and many others. Non-contact enables higher speeds and larger distances. Furthermore, material characterization based on friction does not depend on a semantic label and may have greater utility in applications that interact with surfaces. 

In prior work \cite{Zhang15}, reflectance disks have been used for material recognition. Their work combines material descriptor of textons distribution with similarity preserving binary embedding methods such as 
circulant binary embedding (CBE) \cite{Yu14}, 
bilinear embedding \cite{Gong13a}, iterative quantization (ITQ) \cite{Gong13b},  angular quantization-based binary codes (AQBC) \cite{Gong12}, spectral hashing (SH)  \cite{Weiss09}, locality-sensitive binary codes from shift-invariant kernels (SKLSH)  \cite{Raginsky09}, and locality sensitive hash (LSH) \cite{Charikar02}. 
Traditional methods of texton histograms \cite{Cula01a,Cula01b,Leung01,Varma02} have a weakness in their discriminatory power. 

Fisher Vector pooling as part of deep learning frameworks has been shown to have excellent performance in texture recognition problems \cite{Cimpoi15}. Binary embedding combined with Fisher Vector pooling \cite{perronnin2010large} 
 has been explored for image retrieval. 
Deep learning is known for  discovering discriminant and robust feature representations.  The combination of  deep learning and  binary compact hash codes is  an interesting path that combines the efficiency of binary codes (hamming distance is fast) with the robust performance of deep learning.  Semantic hashing \cite{salakhutdinov2009semantic,torralba2008small} builds a deep binary model by quantizing a pre-trained and fine-tuned stack of Restricted Boltzmann Machines (RBMs). 
Deep hashing \cite{erin2015deep} develops a neural network to learn multiple hierarchical non-linear transformations mapping raw images to compact binary hash codes.
However, these hashing techniques  rely on hand-crafted visual features as input.

Recent work shows that deep convolutional neural network trained on a sufficiently large dataset such as ImageNet \cite{ILSVRC15} can be transferred to other computer vision tasks \cite{gong2014multi,Cimpoi15,donahue2013decaf,razavian2014cnn}.  Lower convolutional layers remain similar on different datasets \cite{zhou2014learning}. 
CNN activations are still sensitive to translation, rotation and scale \cite{gong2014multi}.
Recognizing reflectance disks from unknown illumination and surface tilts is very similar to texture recognition. It is the distribution of features and visual structures, not their particular spatial location, that is most important. Therefore,
dense pooling  methods are essential. 
Multi-scale orderless pooling \cite{gong2014multi} and deep filter banks \cite{Cimpoi15} using VLAD pooling \cite{jegou2010aggregating} and Fisher Vector pooling of CNN activations achieve  state-of-the-art results on texture, material and scene recognition tasks.
We leverage Fisher vector pooling of CNN activations for exploring compact hash codes as binary representations of data.

\begin{figure}[t]
\begin{center}
\includegraphics[width=0.85\linewidth]{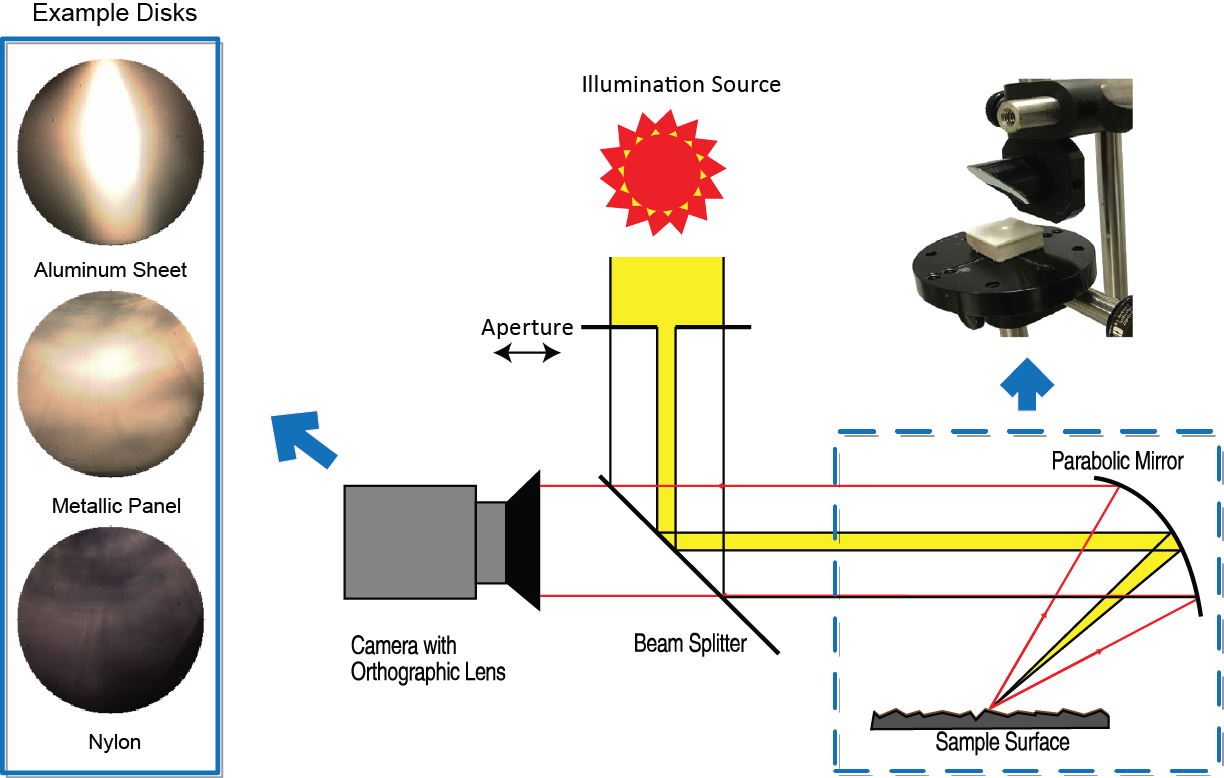}
\end{center}
\caption{ 
Schematic of the mirror-based camera. Reflectance disks are obtained by viewing a single point under multiple viewing directions using a concave parabolic mirror viewed by a telecentric lens. The off-axis concave parabolic mirror is shown in the upper right.}
\label{fig:cameraschematic}
\end{figure}


\section{One-Shot In-Field Reflectance Disks}
\label{sec:reflectance}
We follow Zhang et al.\cite{Zhang15} and use a mirror-based camera \cite{Dana01,Dana04} 
to  measure reflectance of surface points. 
The camera components are an off-axis parabolic mirror, a CCD camera, a movable aperture and a beam splitter (as shown in Figure~\ref{fig:cameraschematic}). 
The parabolic mirror is fixed so that its focus is at the surface point to be measured. The illumination source is a collimated beam of light parallel to the global plane of the surface passing through a movable aperture. The angle of the incident ray  at the surface is determined by the intersection point with the mirror.  Therefore, the illumination 
 direction can be controlled by  planar translations of  the aperture.  Similarly, light reflected  from the surface point is reflected by the mirror to a set of  parallel rays directed through a beam splitter to the camera.  Each pixel of this camera image corresponds to a viewing direction of the surface point.  Therefore, the recorded image, referred to as a {\it reflectance disk}, is an observation of a single point on the surface but from a dense sampling of viewing directions.

\section{Deep Reflectance Codes}



Reflectance naturally encodes the intrinsic physical properties of the surface, but its measurement is sensitive to illumination changes. 
For reflectance disks,  the  illumination changes correspond to translational shifts of image features; e.g. the specularity position shifts. 
We explore translation invariant representation to extract the physical information. Pre-trained CNN features have been applied to a number of computer vision tasks with great success \cite{krizhevsky2012imagenet,jia2014caffe,Lecun15}, but CNN features are 
fairly sensitive to translation and rotation. Dense pooling of CNN activations removes the globally spatial information and therefore is robust to pattern translation and rotation. VLAD-CNN and FV-CNN achieved the state-of-the-art results in scene understanding and material recognition  \cite{gong2014multi,Cimpoi15}.
We leverage the shift-invariance representation of dense pooling CNN for exploring compact hash codes. The resulting hash codes potentially preserve the similarities of surface physical characteristics, which we refer to as {\it deep reflectance codes}.
A key difference from Cimpoi {\it et al.} \cite{Cimpoi15} is that we introduce a unsupervised hashing approach rather than a material descriptor for supervised SVM material classification.

\begin{figure*}[t]
\subfloat[t-SNE of raw image data.]
{
\includegraphics[height=0.32\linewidth]{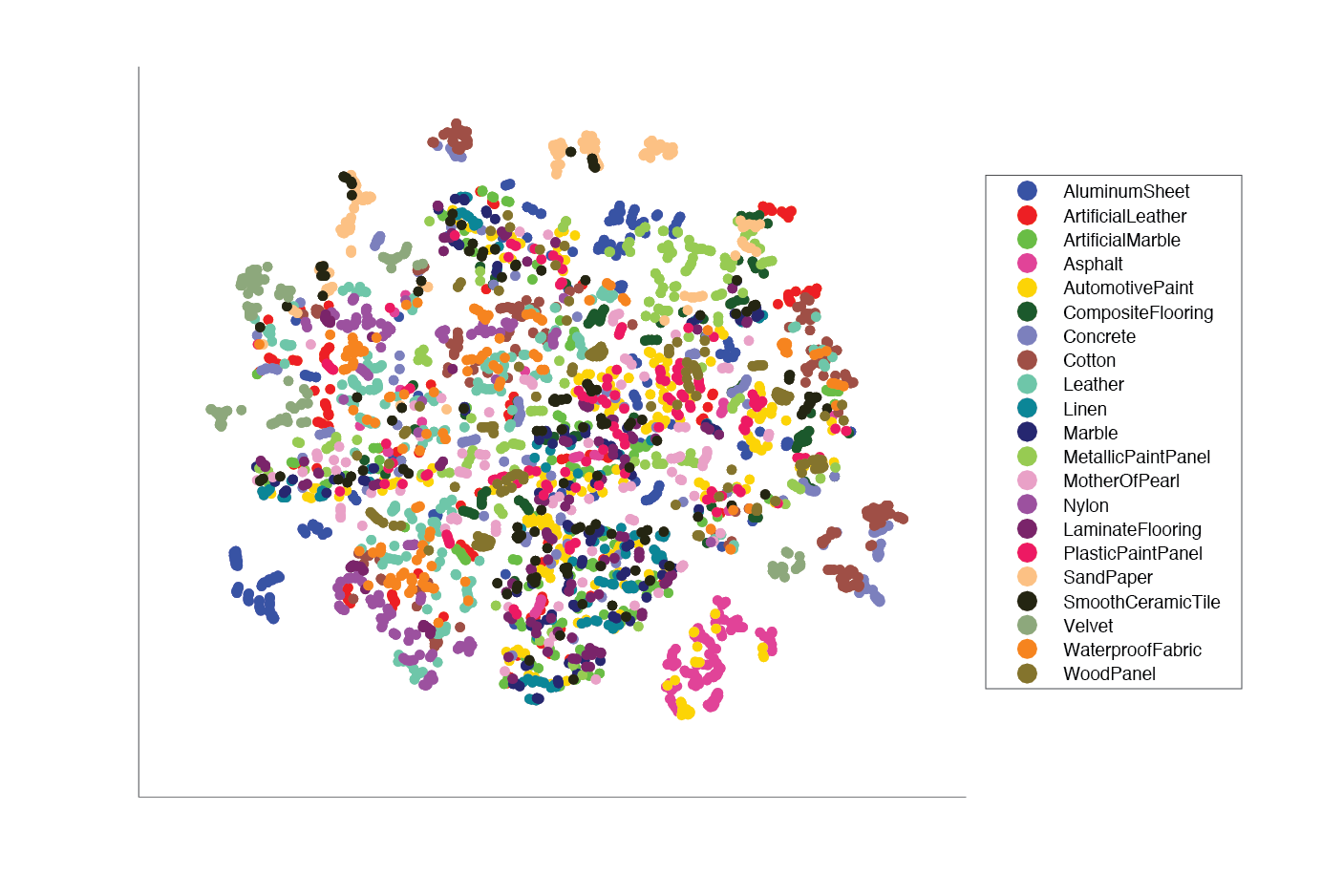}
}
\subfloat[t-SNE of Texton Maps.]
{
\includegraphics[height=0.32\linewidth]{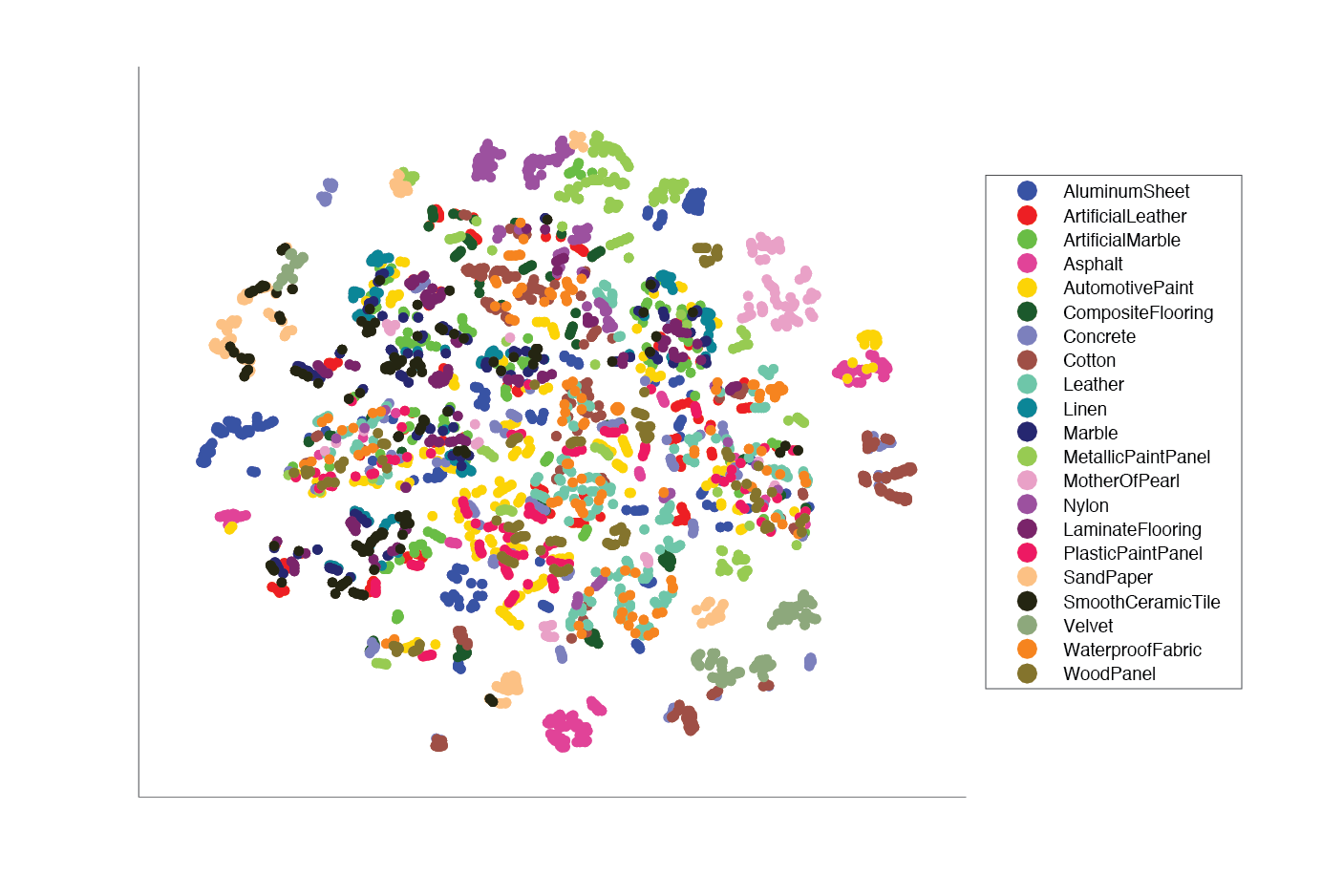}
}\\
\subfloat[Material Manifold of t-SNE FV-CNN. ]
{
\includegraphics[height=0.32\linewidth]{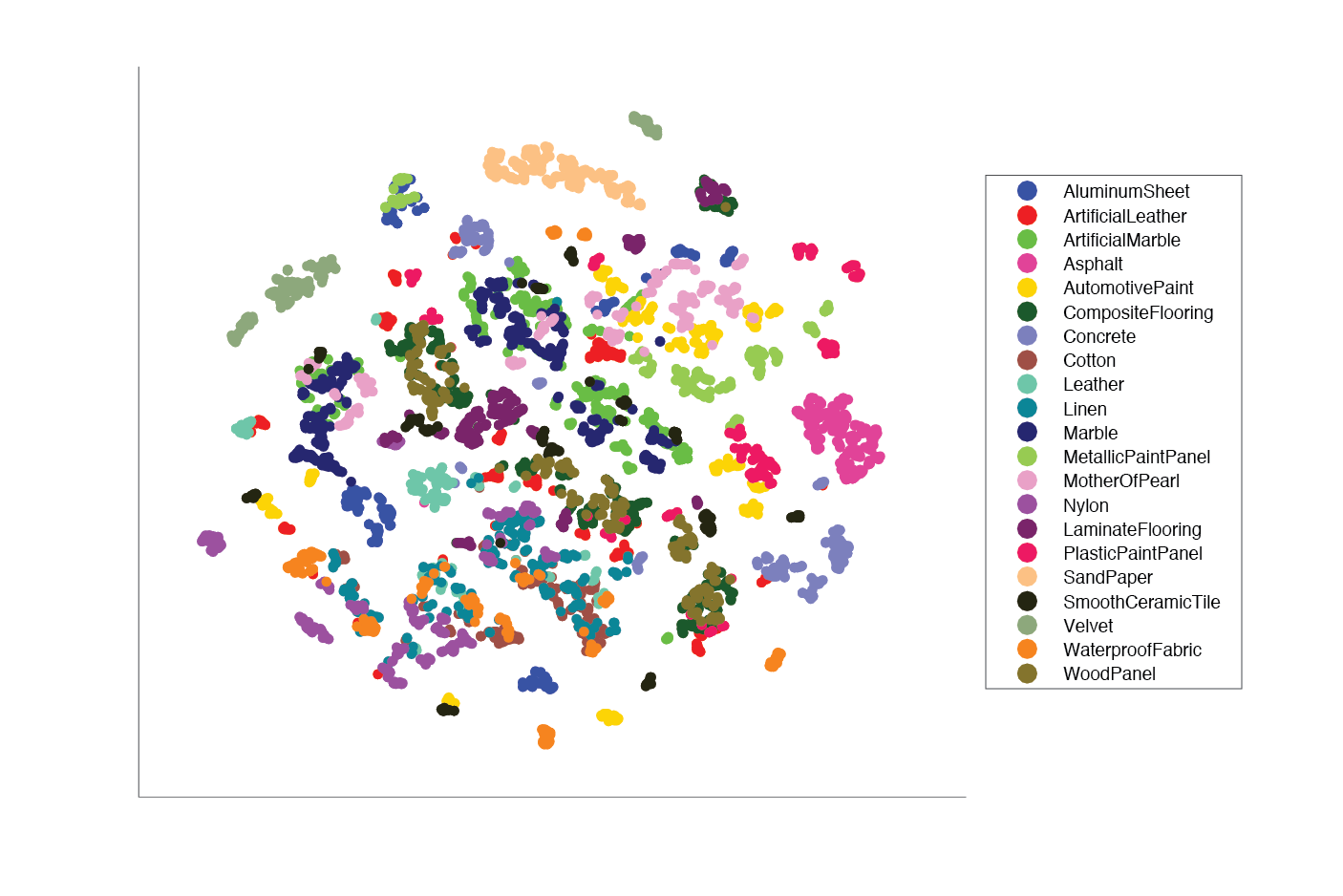}
}
\subfloat[Friction Manifold in t-SNE space.]
{
\includegraphics[height=0.32\linewidth]{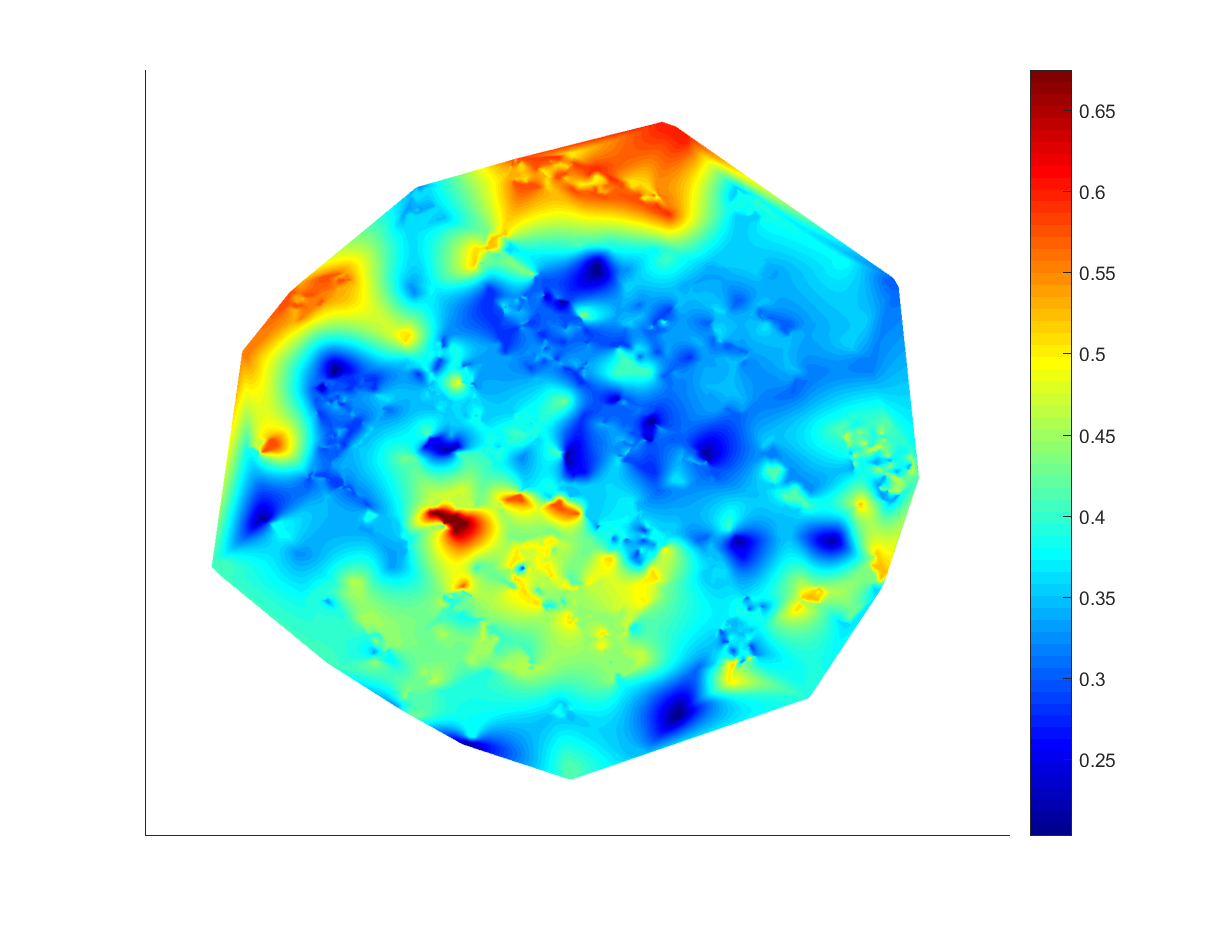}
}
\caption{ (a) and (b) shows t-SNE of raw image data and traditional texton representation. (c) t-SNE embedding of deep reflectance codes representation. Classes are color-coded (21 classes). There are 5-10 instances per class (137 material surfaces). For some classes there is significant intra-class reflectance variation, but most group well within the t-SNE manifold.  (d) Friction map generated in the t-SNE space. Each instance can have a different friction value, as shown in Table~\ref{fig:frictionsensor}. }
\label{fig:embedding}
\end{figure*}

Visualizing the effect of the FV-CNN representation can be done with t-SNE \cite{van2008visualizing}.  Figure~\ref{fig:embedding} shows reflectance disk representations using (a) raw reflectance disk data  and (b) texton maps. The representation of FV-CNN in Figure~\ref{fig:embedding} (c) shows significantly better class discrimination. 
This figure depicts many material instances under multiple illumination directions and 
we can see that our reflectance descriptor  provides sufficient invariance and allows grouping of similar materials.

The challenge of using FV-CNN for retrieval is the high dimensionality of the Fisher vector representation ($64K$ dimensions for our implementation described in Section \ref{sec:ex:hashing}). 
To achieve a more compact representation,  we integrate a binary embedding strategy which preserves the similarity of the Fisher vectors.  The resulting hash codes are suitable for both material recognition and friction estimation. In this work, the recognition experiments are used to evaluate the quality of the representation in order to use it for friction estimation. 


\paragraph{\bf DRC} 

For binary embedding, projection to a lower dimensional subspace that preserves the similarity is a key component. 
The Johnson-Lindenstrauss (J-L) Lemma implies that with high probability, the relative distances between all pair of points are approximately preserved under random projection \cite{arriaga1999algorithmic}. Therefore, by randomly projecting the high-dimensional data into lower dimension, we can still achieve a comparative results with a linear classifier.
Additionally, learning a kernel classifier using Fisher kernel is equivalent to learning a linear classifier on the Fisher vector \cite{perronnin2010improving}. By quantizing the randomly projected Fisher vector to binary hash codes, we are able to approximate the behavior of linear classifier using Fisher Vectors,  approximating a kernel classifier using Fisher Kernel. 
The dot product has been shown as a good measurement of similarity for Fisher vectors and we can use Local Sensitivity Hashing (LSH) \cite{Charikar02} to binarize the Fisher Vector as in \cite{perronnin2010large} quantizing the randomly projected data using the hyper-planes across the origin.  
The cosine similarities of all pairs of data are preserved under random projection \cite{shi2012margin}.
Therefore the hamming distance of generated hash codes preserves the similarity of Fisher vectors. We use quantized random projection of Fisher representation in one variation of our approach, and we refer to this method as 
{\it DRC - Deep Reflectance Codes.}

\paragraph{\bf DRC-opt}
LSH directly quantizes the randomly projected data into hash codes.   Let $v\in \mathbb{R}^d$ represent a projected data point and the hash function $\mathrm{sgn}(v)$ maps the data to the vertex of the hyper-cube $\{-1,1\}^d$. The quantization error can be written as $\|\mathrm{sgn}(v)-v\|^2$. Following the iterative quantization method \cite{Gong13b}, in order to preserve the local similarity of the data, a better hash code can be learned by rotating the data to minimize the quantization error:
\begin{equation}
\sum_{i=1}^N\|b_i-v_iR\|_2^2,
\end{equation}
where $N$ is the number of training data, the binary code is given by $b_i=\mathrm{sgn}(v_i R)$, $R\in \mathbb{R}^{d\times d}$ is a rotation matrix.
We use this combination of fisher vector CNN with this optimized binary embedding and we refer to it as 
{\it DRC-opt optimized Deep Reflectance Codes}.

\begin{table*}[t]
\centering
\begin{adjustbox}{width=\textwidth}
\rowcolors{2}{TableOdd}{TableEven}
\begin{tabular}{
r!{\color{TableBorder}\VRule[0.5pt]}
c!{\color{TableBorder}\VRule[0.5pt]}
c!{\color{TableBorder}\VRule[0.5pt]}
c!{\color{TableBorder}\VRule[0.5pt]}
c!{\color{TableBorder}\VRule[0.5pt]}
c!{\color{TableBorder}\VRule[0.5pt]}
c!{\color{TableBorder}\VRule[0.5pt]}
c!{\color{TableBorder}\VRule[0.5pt]}
c!{\color{TableBorder}\VRule[0.5pt]}
c!{\color{TableBorder}\VRule[0.5pt]}
c!{\color{TableBorder}\VRule[0.5pt]}
c!{\color{TableBorder}\VRule[0.5pt]}
c!{\color{TableBorder}\VRule[0.5pt]}
}
\arrayrulecolor{TableBorder}

\rowcolor{TableOdd}
\TextBorder{Friction} 
&\CellLightGray{\TextGreen{AVG.}} & \TextGreen{STD.} & \CellLightGray{\TextGreen{1}} & \TextGreen{2} & \CellLightGray{\TextGreen{3}} & \TextGreen{4} & \CellLightGray{\TextGreen{5}} & \TextGreen{6} & \CellLightGray{\TextGreen{7}} & \TextGreen{8} & \CellLightGray{\TextGreen{9}} & \TextGreen{10}\\
\specialrule{0.5pt}{0pt}{0pt}

\TextRed{AluminumSheet}&\TextBorder{0.354} & \TextBorder{0.018} & 0.354 & 0.344 & 0.364 & 0.384 & 0.344 & 0.335 &  &  &  & \\
\specialrule{0.5pt}{0pt}{0pt} 
\TextRed{ArtificialLeather}&\CellLightGray{\TextBorder{0.463}} & \TextBorder{0.099} & \CellLightGray{0.325} & 0.510 & \CellLightGray{0.577} & 0.404 & \CellLightGray{0.499} &  & \CellLightGray{ }&  & \CellLightGray{ }& \\
\specialrule{0.5pt}{0pt}{0pt} 
\TextRed{ArtificialMarble}&\TextBorder{0.296} & \TextBorder{0.013} & 0.287 & 0.306 & 0.306 & 0.306 & 0.296 & 0.296 & 0.315 & 0.296 & 0.277 & 0.277  \\
\specialrule{0.5pt}{0pt}{0pt} 
\TextRed{Asphalt}&\CellLightGray{\TextBorder{0.398}} & \TextBorder{0.043} & \CellLightGray{0.384} & 0.404 & \CellLightGray{0.414} & 0.335 & \CellLightGray{0.466} & 0.384 & \CellLightGray{ }&  & \CellLightGray{ }& \\
\specialrule{0.5pt}{0pt}{0pt} 
\TextRed{AutomotivePaint}&\TextBorder{0.342} & \TextBorder{0.019} & 0.344 & 0.306 & 0.335 & 0.344 & 0.335 & 0.354 & 0.344 & 0.374 &  & \\
\specialrule{0.5pt}{0pt}{0pt} 
\TextRed{CompositeFlooring}&\CellLightGray{\TextBorder{0.371}} & \TextBorder{0.016} & \CellLightGray{0.384} & 0.374 & \CellLightGray{0.384} & 0.354 & \CellLightGray{0.344} & 0.384 & \CellLightGray{0.384} & 0.354 & \CellLightGray{0.384} & 0.364  \\
\specialrule{0.5pt}{0pt}{0pt} 
\TextRed{Concrete}&\TextBorder{0.453} & \TextBorder{0.060} & 0.424 & 0.364 & 0.435 & 0.456 & 0.521 & 0.521 &  &  &  & \\
\specialrule{0.5pt}{0pt}{0pt} 
\TextRed{Cotton}&\CellLightGray{\TextBorder{0.449}} & \TextBorder{0.006} & \CellLightGray{0.445} & 0.445 & \CellLightGray{0.456} & 0.445 & \CellLightGray{0.456} &  & \CellLightGray{ }&  & \CellLightGray{ }& \\
\specialrule{0.5pt}{0pt}{0pt} 
\TextRed{Leather}&\TextBorder{0.494} & \TextBorder{0.108} & 0.466 & 0.477 & 0.466 & 0.675 & 0.384 &  &  &  &  & \\
\specialrule{0.5pt}{0pt}{0pt} 
\TextRed{Linen}&\CellLightGray{\TextBorder{0.467}} & \TextBorder{0.031} & \CellLightGray{0.499} & 0.510 & \CellLightGray{0.466} & 0.435 & \CellLightGray{0.445} & 0.445 & \CellLightGray{ }&  & \CellLightGray{ }& \\
\specialrule{0.5pt}{0pt}{0pt} 
\TextRed{Marble}&\TextBorder{0.320} & \TextBorder{0.028} & 0.344 & 0.306 & 0.335 & 0.325 & 0.277 & 0.287 & 0.296 & 0.344 & 0.364 & 0.325  \\
\specialrule{0.5pt}{0pt}{0pt} 
\TextRed{MetallicPaintPanel}&\CellLightGray{\TextBorder{0.331}} & \TextBorder{0.018} & \CellLightGray{0.344} & 0.335 & \CellLightGray{0.335} & 0.315 & \CellLightGray{0.306} & 0.354 & \CellLightGray{ }&  & \CellLightGray{ }& \\
\specialrule{0.5pt}{0pt}{0pt} 
\TextRed{MotherOfPearl}&\TextBorder{0.333} & \TextBorder{0.023} & 0.306 & 0.306 & 0.344 & 0.344 & 0.335 & 0.364 &  &  &  & \\
\specialrule{0.5pt}{0pt}{0pt} 
\TextRed{Nylon}&\CellLightGray{\TextBorder{0.412}} & \TextBorder{0.015} & \CellLightGray{0.435} & 0.424 & \CellLightGray{0.424} & 0.394 & \CellLightGray{0.404} & 0.404 & \CellLightGray{0.414} & 0.394 & \CellLightGray{ }& \\
\specialrule{0.5pt}{0pt}{0pt} 
\TextRed{LaminateFlooring}&\TextBorder{0.394} & \TextBorder{0.018} & 0.394 & 0.384 & 0.374 & 0.384 & 0.404 & 0.424 & & & &
 \\
\specialrule{0.5pt}{0pt}{0pt} 
\TextRed{PlasticPaintPanel}&\CellLightGray{\TextBorder{0.343}} & \TextBorder{0.042} & \CellLightGray{0.325} & 0.325 & \CellLightGray{0.414} & 0.315 & \CellLightGray{0.374} & 0.306 & \CellLightGray{ }&  & \CellLightGray{ }& \\
\specialrule{0.5pt}{0pt}{0pt} 
\TextRed{SandPaper}&\TextBorder{0.547} & \TextBorder{0.039} & 0.510 & 0.532 & 0.499 & 0.577 & 0.577 & 0.589 &  &  &  & \\
\specialrule{0.5pt}{0pt}{0pt} 
\TextRed{SmoothCeramicTile}&\CellLightGray{\TextBorder{0.224}} & \TextBorder{0.021} & \CellLightGray{0.259} & 0.203 & \CellLightGray{0.222} & 0.213 & \CellLightGray{0.222} &  & \CellLightGray{ }&  & \CellLightGray{ }& \\
\specialrule{0.5pt}{0pt}{0pt} 
\TextRed{Velvet}&\TextBorder{0.564} & \TextBorder{0.026} & 0.601 & 0.532 & 0.554 & 0.577 & 0.554 &  &  &  &  & \\
\specialrule{0.5pt}{0pt}{0pt} 
\TextRed{WaterproofFabric}&\CellLightGray{\TextBorder{0.394}} & \TextBorder{0.014} & \CellLightGray{0.404} & 0.414 & \CellLightGray{0.394} & 0.394 & \CellLightGray{0.384} & 0.374 & \CellLightGray{ }&  & \CellLightGray{ }& \\
\specialrule{0.5pt}{0pt}{0pt} 
\TextRed{WoodPanel}&\TextBorder{0.346} & \TextBorder{0.031} & 0.364 & 0.354 & 0.354 & 0.344 & 0.374 & 0.287 & 
\multicolumn{4}{c}{\multirow{7}{*}[0.95in]{\cellcolor{white}
\includegraphics[width=1.3in,height=0.95in]{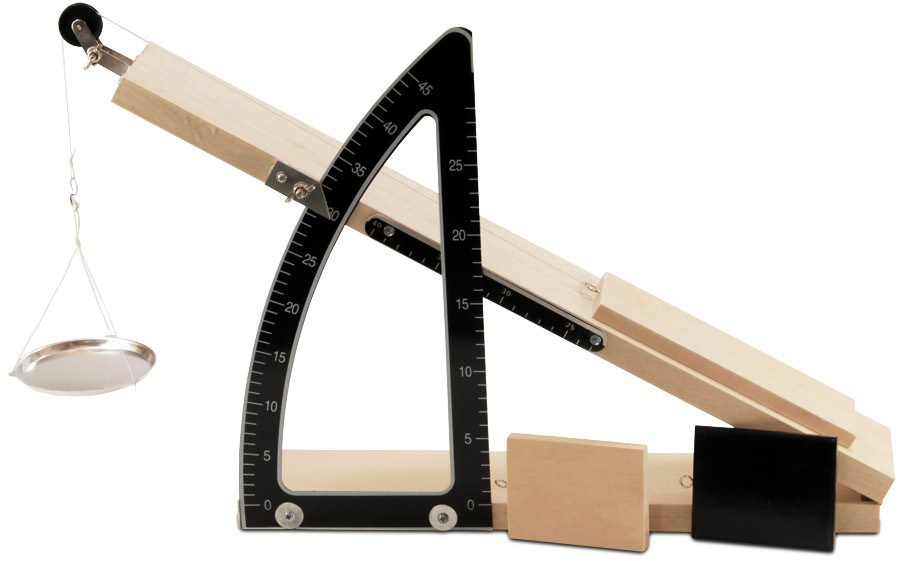}
}} \\
\specialrule{0.5pt}{0pt}{0pt} 


\end{tabular}
\end{adjustbox}
\caption{The measured friction coefficients of the material dataset; the friction sensor is shown at the right bottom corner.
The collection of 137 material surfaces are grouped into classes. Each class has multiple distinct instances numbered from 1 to 10.}
\label{fig:frictionsensor}
\end{table*}

\section{Friction from Reflectance}
\label{sec:friction}

Deep reflectance codes give us a compact representation that encodes the reflectance disks which we expect to encode rich information about the physical properties of the material itself. We are particularly interested in probing tactile properties of a surface from its reflectance appearance. For this, we focus on estimating friction from reflectance disks. In this section, we introduce the first-of-its-kind friction-reflectance database and then apply deep reflectance codes to estimate friction from one-shot in-field reflectance disks.


\subsection{Friction-Reflectance database}
\label{sec:dataset}

We collect 137 different materials (shown in Figure \ref{fig:database}) and group them into 21 categories: 
aluminum sheet, artificial leather, artificial marble, asphalt, automotive paint, composite flooring, concrete, cotton, leather, linen, marble, metallic paint panel, mother of pearl, nylon,  painted wood flooring, plastic paint panel, sand paper, smooth ceramic tile, velvet, waterproof fabric, wood panel. 
Our database includes 5-10 different material samples (instances)  per category,  and we measure 2 different surface spots per sample  with 7 illumination directions $(-20^\circ,-10^\circ,0,10^\circ,20^\circ)$  along axis and $(-10^\circ,10^\circ)$ off axis,
where $0^\circ$ is frontal illumination aligned with the surface normal. Additionally, images with 3 exposure settings are collected for high dynamic range imaging. The total number of reflectance disks are 5754. 

Friction of a surface is an intrinsic physical property of a surface that can readily be measured non-visually with a contact device. The coefficient of kinetic friction multiplied by the force normal (due to gravity) is the force of friction which act to hold an object in place on a surface.
We adopt a simple approach for measuring kinetic friction coefficient by using an inclined plane.  In the experiments, we hold the object just above the surface of the inclined plane and release it. At low angles, the object should does not move. As the angle increases,  the object begins to slide down at a constant velocity. The angle $\theta$ is recorded, and the kinetic friction coefficient is calculated as $\mu=\tan\theta$. For a thin material such as cloth, we attach it to a solid object for the friction measurement. The measured kinetic friction coefficients for all 137 surfaces  and the inclined plane friction measurement device are shown in the Table \ref{fig:frictionsensor}. 
The measured kinetic friction coefficients are shown in Table \ref{fig:frictionsensor}. 

\subsection{Hashing for Friction Prediction}

Deep reflectance codes encode rich information of physical properties of material surface from reflectance disks. 
The similarities of material physical properties are preserved in Hamming distance by compact hash codes. 
Figure \ref{fig:embedding} (c,d) visualizes the correspondence between the friction distribution and the representation using t-SNE, and we are able to see the material surfaces with similar friction coefficients are likely to have small distance. 
Therefore, our approach of deep reflectance codes 
can be used for friction estimation. 
We build a binary hash table with the DRC as the key and the corresponding surface friction coefficient as the hash value. 
The binary representations are compact and amenable for fast nearest neighbor retrieval. 
The predicted friction coefficient of a sample is given by the average of friction coefficients of retrieved samples.

\section{Experimental Results}

To evaluate the proposed approach of deep reflectance codes in spatial invariance representation and friction prediction, we conduct two groups of experiments. We first apply deep reflectance codes on retrieval for material recognition as a litmus test of the representation and then evaluate the performance of friction prediction from one-shot measurements. 


\subsection{Hashing for Material Recognition}
\label{sec:ex:hashing}

Material and texture are typically modeled by orderless pooling of visual patterns, {\it i.e.} texton distributions, as the shape and spatial information are not important. We apply deep reflectance codes on image retrieval for material recognition to demonstrate the discriminative power and the spatial invariance of our representation. 

\paragraph{\bf Additional Datasets} 
We also consider three other material datasets to evaluate our DRC approach for general material recognition. The first one is Flickr Material Dataset  (FMD) \cite{Sharan13}, which contains 10 material classes. The second one is the  Describable Texture Datasets (DTD) \cite{Cimpoi14}, containing texture images labeled with 47 describable attributes. The third texture dataset is KTH-TIPS-2b (KTH) \cite{caputo2005class}, which contains 11 material categories with 4 samples per category and a number of example images for each sample. For this unsupervised hashing for material recognition experiments, the training set means the retrieval set for nearest neighbor search. For the DTD and FMD, the test set are given by randomly selected $20\%$ images, and the rest are for training.  The test set of reflectance database and KTH-TIPS are given by randomly selecting one sample of each material category, and the remaining samples are used for training. The experimental results are averaged in 10 runs. 

\begin{table*}[t]
\centering
\begin{adjustbox}{width=0.8\textwidth}
\newcolumntype{C}[1]{>{\centering\let\newline\\\arraybackslash\hspace{0pt}}m{#1}}

\rowcolors{2}{TableOdd}{TableEven}
\begin{tabular}{
r!{\color{TableBorder}\VRule[0.5pt]}
c!{\color{TableBorder}\VRule[0.5pt]}
c!{\color{TableBorder}\VRule[0.5pt]}
c!{\color{TableBorder}\VRule[0.5pt]}
c!{\color{TableBorder}\VRule[0.5pt]}
c!{\color{TableBorder}\VRule[0.5pt]}
c!{\color{TableBorder}\VRule[0.5pt]}
c!{\color{TableBorder}\VRule[0.5pt]}
}
\arrayrulecolor{TableBorder}
\rowcolor{TableOdd}

\TextBorder{dataset} 
& \CellLightGray{\TextGreen{FV-SIFT-Hash}}& 
\TextGreen{  CNN-ITQ  } & 
\CellLightGray{\TextGreen{VLAD-CNN-KBE}} &
\TextGreen{FV-CNN-KBE}&
\CellLightGray{\TextGreen{     DRC    }}&
\TextGreen{   DRC-opt   }\\
\specialrule{0.5pt}{0pt}{0pt}


\TextRed{reflectance}&\TextBorder{64.5{\tiny $\pm 5.8$}} & 51.9{\tiny $\pm 7.0$} & 60.1{\tiny $\pm 6.7$} & 58.8{\tiny $\pm 7.1$} & 59.9{\tiny $\pm 4.8$} & 60.2{\tiny $\pm 5.1$}  \\
\specialrule{0.5pt}{0pt}{0pt} 
\TextRed{FMD}&\CellLightGray{48.3{\tiny $\pm 4.5$}} & 65.0{\tiny $\pm 1.7$} & \CellLightGray{59.4{\tiny $\pm 2.8$}} & 57.7{\tiny $\pm 2.7$} & \CellLightGray{64.8{\tiny $\pm 2.8$}} & \TextBorder{65.5{\tiny $\pm 3.0$}}  \\
\specialrule{0.5pt}{0pt}{0pt} 
\TextRed{DTD}&43.6{\tiny $\pm 1.4$} & 52.6{\tiny $\pm 1.3$} & 52.3{\tiny $\pm 1.4$} & 53.1{\tiny $\pm 0.9$} & 55.4{\tiny $\pm 1.2$} & \TextBorder{55.8{\tiny $\pm 1.4$}}  \\
\specialrule{0.5pt}{0pt}{0pt} 
\TextRed{KTH}&\CellLightGray{72.0{\tiny $\pm 4.3$}} & 73.7{\tiny $\pm 3.4$} & \CellLightGray{75.6{\tiny $\pm 6.2$}} & 74.4{\tiny $\pm 5.2$} & \CellLightGray{76.6{\tiny $\pm 5.2$}} & \TextBorder{77.2{\tiny $\pm 6.4$}}  \\
\specialrule{0.5pt}{0pt}{0pt} 

\end{tabular}
\end{adjustbox}
\vspace{0.2in}
\caption{Comparison of encodings and hashing methods (using $1024$-bit hash codes) on different material datasets. The recognition precision over 10 runs, large standard deviation of reflectance and KTH dataset due to randomly selecting one sample surface from each class as test set in each iteration (described in Section \ref{sec:ex:hashing}).}
\label{tab:hash}
\end{table*}

\paragraph{\bf Baseline Methods}
For general evaluation of our DRC in material recognition, we compare it with the following methods and settings: FV-SIFT-Hash \cite{perronnin2010large}, CNN-ITQ \cite{Gong13b}, VLAD-CNN-KBE \cite{zhangfast}, FV-CNN-KBE. 
FV-SIFT-Hash is binarized fisher vector pooling of SIFT feature as in \cite{perronnin2010large}. CNN-ITQ is iterative quantization binary hashing \cite{Gong13b} on $4096$-dimensional CNN activations, which is typically compared as a baseline for deep hashing algorithm. KBE represents a latest work Kronecker Binary Embedding \cite{zhangfast} (randomized orthogonal version with order $2$ is used in this experiment, which is good for high-dimension data). 
We define ground truth as the given class label of material, and use the mode of labels from 10 retrieved nearest neighbors of hash codes in Hamming Distance as the predicted label. The mean recognition precision is computed over all inquiries from test set averaged over 10 runs.  

In our experiments, a pre-trained VGG-M model \cite{chatfield2014return} is used for computing CNN activations. CNN activations for VLAD are taken by the output on the first fully connected layer (full6) with a $4096$ dimension. We extract activations for $128\times 128$ paches sampled with a stride of $32$ pixels, and use PCA to reduce them to $512$ dimensions \cite{gong2014multi}. We use 64 $k$-means centers for computing VLAD pooling, the vector is given by assigning each patch to its nearest clustering center and aggregating the residuals of patches minus the center. Hence, the dimension of VLAD pooling CNN is $32K$. The CNN activations for FV pooling are taken from the output of the first convolutional layer with $512$ dimensions (conv5 of VGG-M). We use $64$ Gaussian components for the FV representation, resulting in $65K$ dimensions FV-CNN descriptors. The implementations are based on MatConvNet~\cite{vedaldi15matconvnet} and VLFeat~\cite{vedaldi08vlfeat}.

Table \ref{tab:hash} shows material recognition results comparing different encodings and hashing methods. We can see the randomized and optimized version of DRC hashing outperform the other methods in FTD, DTD and KTH datasets. Note that our compact unsupervised binary hashing methods can achieve comparable results with training SVM on FV-CNN as reported in \cite{Cimpoi14}. FV pooling provides domain transfer of CNN activations and we have found it make the use of the property that FV pooling essentially discarding the influence the CNN activations corresponding to frequent background  structures that are often domain specific. 
DRC-opt ($60.2\%$) is much better than directly using CNN activations (CNN-ITQ $51.9\%$) for the reflectance dataset.  The methods using CNN features are slightly worse than hashing with FV pooling SIFT feature, since the reflectance disks dataset is very different from the CNN pre-trained dataset, ImageNet, which consists real world images. 
Reflectance Hashing with textons \cite{Zhang15} has an overall recognition precision of $56.69\%$ on this reflectance dataset. We  test  this method on the reflectance dataset only since it is specifically developed for reflectance disks. 
Our evaluation indicates the utility of this binary code representation both for general material recognition and specifically for reflectance disk representation. 

\begin{figure*}[t]
\begin{center}
\subfloat
{
\includegraphics[height=0.35\linewidth]{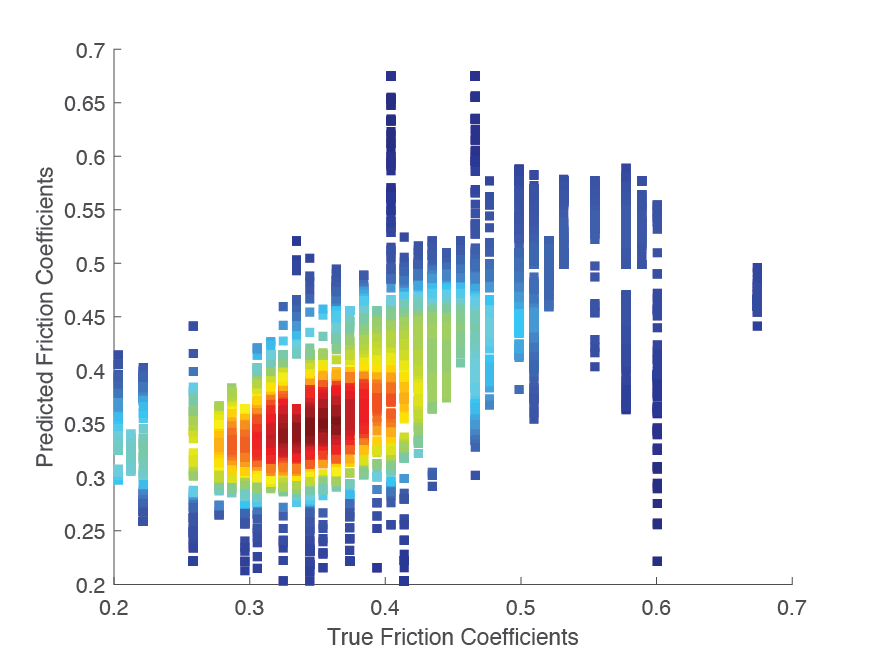}
}
\subfloat
{
\includegraphics[height=0.35\linewidth]{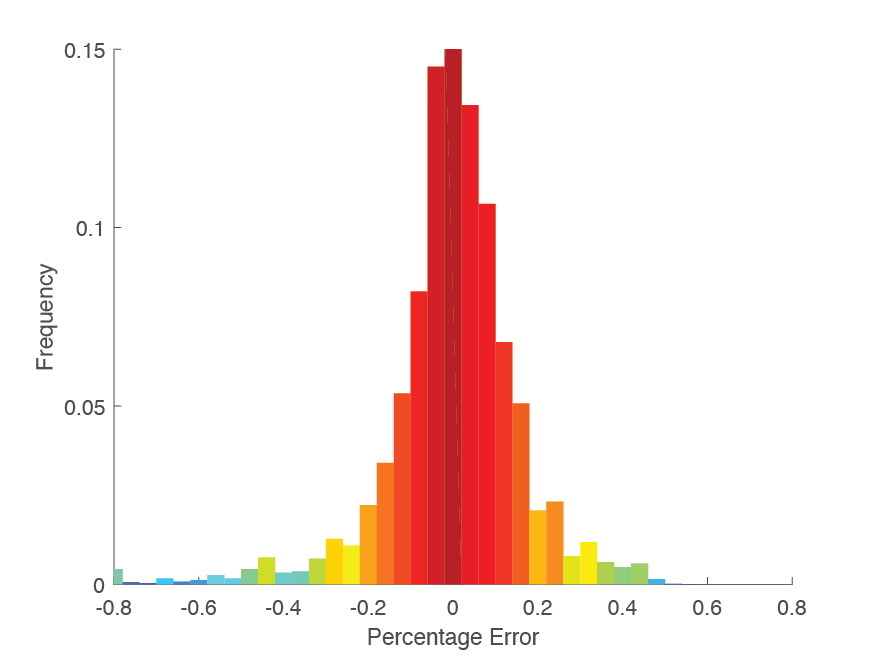}
}\\
\subfloat
{
\includegraphics[height=0.35\linewidth]{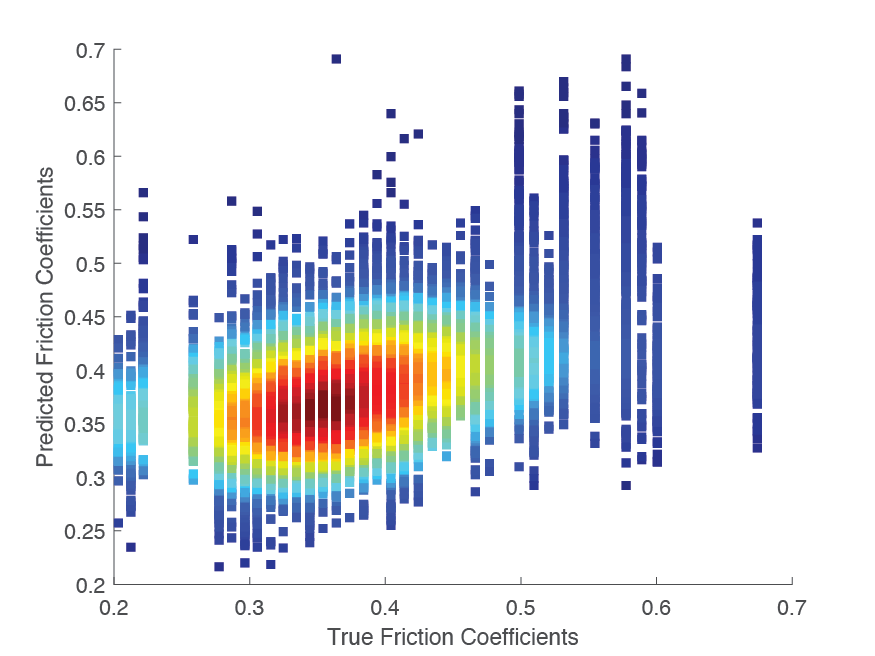}
}
\subfloat
{
\includegraphics[height=0.35\linewidth]{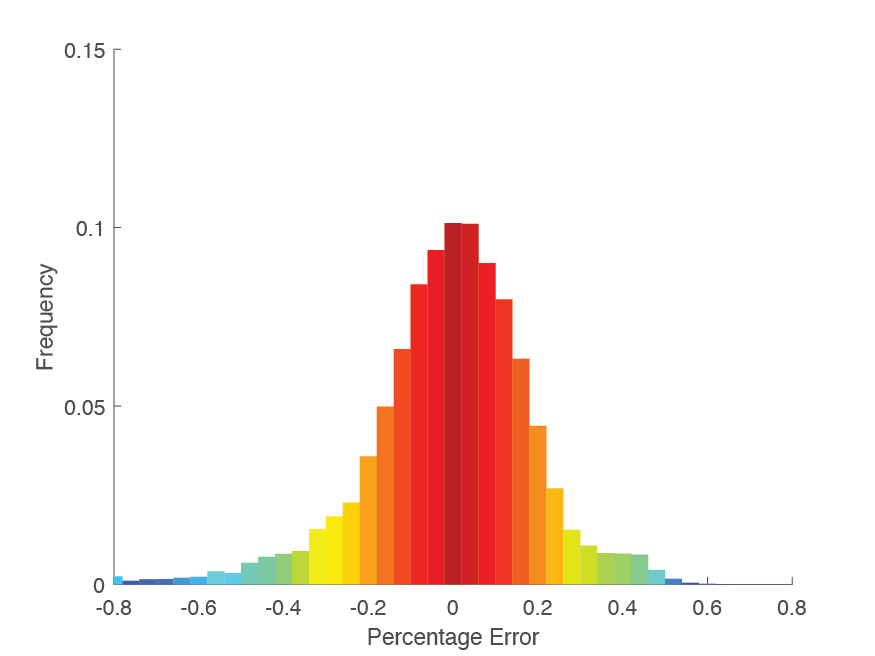}
}
\end{center}
\caption{
(Top-left) Friction prediction scatter plot, colored by density. (Top-right) Percentage error distribution, the average percentage error is $11.15\%$. (Bottom row) The friction prediction with neural net regression has the average percentage error of $14.74\%$. }
\label{fig:scatter}
\end{figure*}

\begin{figure*}[t]
\centering
\begin{adjustbox}{max width=\textwidth}
\includegraphics[width=\linewidth]{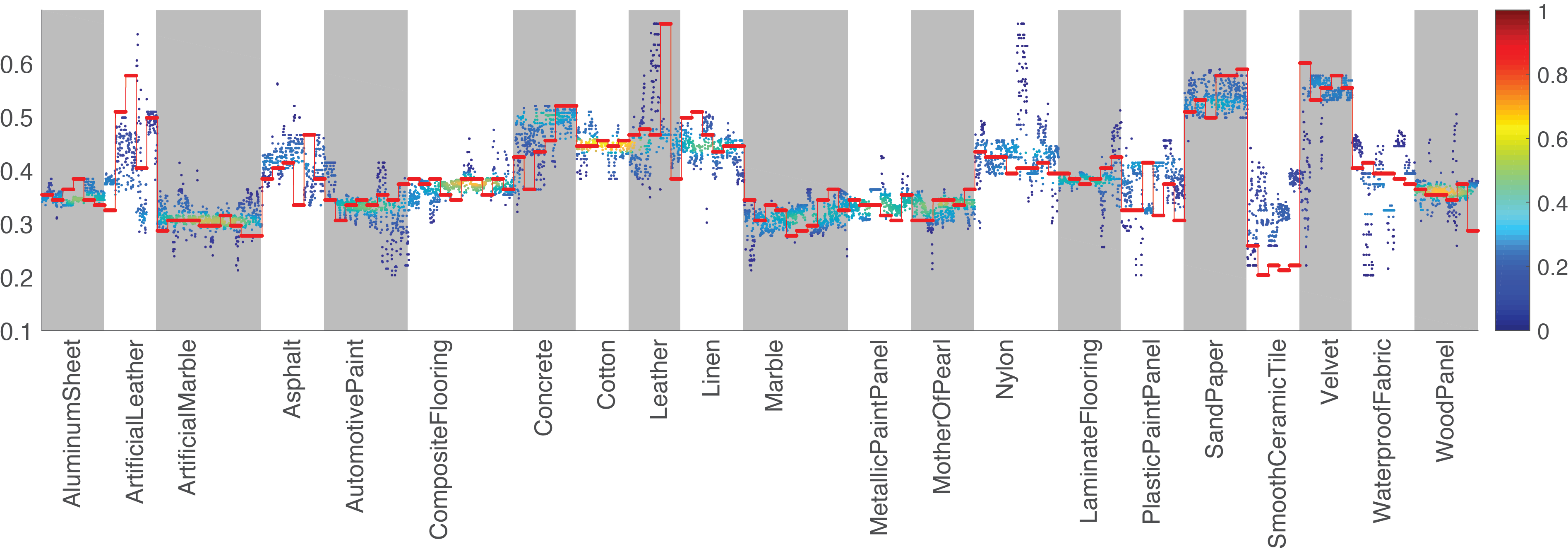}
\end{adjustbox}
\begin{adjustbox}{max width=\textwidth}
\includegraphics[width=\linewidth]{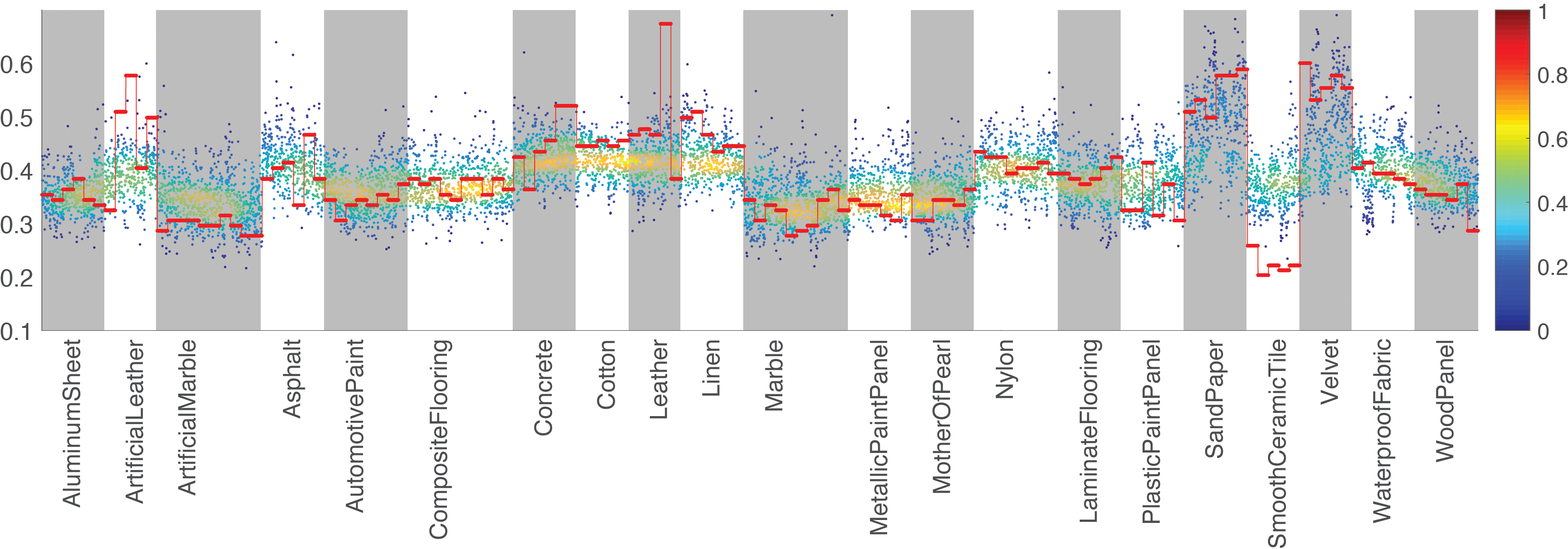}
\end{adjustbox}
\caption{
(Top) Per-class friction prediction results, colored by local density. The red line is the ground truth of friction coefficients. The mean percentage error is $11.15\%$. The predicted coefficients match the overall friction variation with material class. Friction-from-reflectance has never been attempted to our knowledge and these results show great promise for this challenging task. (Bottom) The friction prediction with neural net regression has the average percentage error of $14.74\%$. 
}
\label{fig:density}
\end{figure*}

\subsection{Friction  Prediction}

For friction prediction experiments, we use 1024-bit deep reflectance codes to estimate the surface friction from one-shot in-field reflectance disks.  For reflectance-friction dataset, each surface instance (from same or different material class) can have a different friction coefficient, as shown in Table \ref{fig:frictionsensor}.  
We use leave-one-out cross-validation as training test splits. Each time the disks of one surface is selected as test set and the disks of the rest 136 surfaces are used for training (retrieval). The predicted friction coefficient is given by calculating the average of retrieved 10 neighbors. 


\paragraph{\bf Baseline Method } To evaluate our hash codes on friction prediction, we use a regression network as a baseline method. The feedforward network takes the material reflectance representation to directly predict the friction coefficients 
by  fitting a parameterized functional form using FV-CNN representation as input (projected into 1024 dimension). 
Each neuron in the network nonlinearly transforms  the weighted received inputs into outputs. Our regression net has two hidden layers with 16 and 8 neurons respectively and apply a sigmoid function as activation function.  The optimal values for parameters are learned by minimizing the sum of the square errors between measured friction coefficients and those predicted by the network.

Figure \ref{fig:scatter} shows the prediction results of DRC and the baseline method. 
The scatter plot (top-left) is color coded and  the colors represent the local density around each point. The  plot shows the largest density of points have low prediction error.
 The top-right figure shows a histogram of the error percentage. The predicted data mainly fall into small error bins
 and the overall mean percentage error is $11.15\%$. The results of baseline method are shown at the bottom and the overall mean percentage error is $14.74\%$.   
 
Figure \ref{fig:density} provides a more detailed view of the friction-from-reflectance prediction by showing the  per-class recognition performance. The red line is the ground truth of friction coefficients, and the points are predicted values colored by the local density of points. Our DRC approach outperforms the baseline method. 
However, the DRC as a hashing approach has the limitation for predicting the surface with extremely high or low friction  due to absence of certain data from retrieval set. 

 


\section{Conclusions}

We have presented the first-of-its kind friction-from-reflectance framework.  Using a unique method for capturing one-shot reflectance, we obtain a simple snapshot of local reflectance. We implement a representation of reflectance  with the goal of using this representation for physical property inference.  As an evaluation of the quality of the representation, we consider its use in retrieval of reflectance disks and also in general image retrieval; a representation that is sufficiently descriptive for recognition is a good candidate 
for input to our novel image-based friction prediction.  Friction-from-prediction will have good practical value in applications including
automated assessment of road conditions prior to vehicular contact, planning robotic grasping of unknown objects, attribute tagging in scene analysis. The paradigm of inferring physical surface properties from image data is the  major contribution of the work. 
The database of reflectance disks, surface images, and corresponding friction measurements will be made publicly available for future research.

\section*{Acknowledgment}
This work was supported by National Science Foundation award
IIS-1421134 to KD and KN and
Office of Naval Research grant N00014-16-1-2158 (N00014-14-1-0316) to KN.
\clearpage

\bibliographystyle{splncs03}
\bibliography{eccv16}

\begin{thebibliography}{10}
\providecommand{\url}[1]{\texttt{#1}}
\providecommand{\urlprefix}{URL }

\bibitem{arriaga1999algorithmic}
Arriaga, R., Vempala, S., et~al.: An algorithmic theory of learning: Robust
  concepts and random projection. In: Foundations of Computer Science, 1999.
  40th Annual Symposium on. pp. 616--623. IEEE (1999)

\bibitem{bell15minc}
Bell, S., Upchurch, P., Snavely, N., Bala, K.: Material recognition in the wild
  with the materials in context database. Computer Vision and Pattern
  Recognition (CVPR)  (2015)

\bibitem{bertaux2007relationship}
Bertaux, E., Lewandowski, M., Derler, S.: Relationship between friction and
  tactile properties for woven and knitted fabrics. Textile Research Journal
  77(6),  387--396 (2007)

\bibitem{caputo2005class}
Caputo, B., Hayman, E., Mallikarjuna, P.: Class-specific material
  categorisation. In: Computer Vision, 2005. ICCV 2005. Tenth IEEE
  International Conference on. vol.~2, pp. 1597--1604. IEEE (2005)

\bibitem{Charikar02}
Charikar, M.S.: Similarity estimation techniques from rounding algorithms. In:
  Proceedings of the thiry-fourth annual ACM symposium on Theory of computing.
  pp. 380--388. ACM (2002)

\bibitem{chatfield2014return}
Chatfield, K., Simonyan, K., Vedaldi, A., Zisserman, A.: Return of the devil in
  the details: Delving deep into convolutional nets. In: British Machine Vision
  Conference (2014)

\bibitem{Cimpoi14}
Cimpoi, M., Maji, S., Kokkinos, I., Mohamed, S., Vedaldi, A.: Describing
  textures in the wild. In: Computer Vision and Pattern Recognition (CVPR),
  2014 IEEE Conference on. pp. 3606--3613. IEEE (2014)

\bibitem{Cimpoi15}
Cimpoi, M., Maji, S., Vedaldi, A.: Deep filter banks for texture recognition
  and segmentation. In: Proceedings of the IEEE Conference on Computer Vision
  and Pattern Recognition. pp. 3828--3836 (2015)

\bibitem{Cula01b}
Cula, O.G., Dana, K.J.: Compact representation of bidirectional texture
  functions. IEEE Conference on Computer Vision and Pattern Recognition  1,
  1041--1067 (December 2001)

\bibitem{Cula01a}
Cula, O.G., Dana, K.J.: Recognition methods for 3d textured surfaces.
  Proceedings of SPIE Conference on Human Vision and Electronic Imaging VI
  4299,  209--220 (January 2001)

\bibitem{Dana01}
Dana, K.J.: Brdf/btf measurement device. International Conference on Computer
  Vision  2,  460--6 (July 2001)

\bibitem{Dana04}
Dana, K., Wang, J.: Device for convenient measurement of spatially varying
  bidirectional reflectance. Journal of the Optical Society of America A  21,
  pp. 1--12 (January 2004)

\bibitem{donahue2013decaf}
Donahue, J., Jia, Y., Vinyals, O., Hoffman, J., Zhang, N., Tzeng, E., Darrell,
  T.: Decaf: A deep convolutional activation feature for generic visual
  recognition. In: Proceedings of The 31st International Conference on Machine
  Learning. pp. 647--655 (2014)

\bibitem{Erdogan11}
Erdogan, G., Alexander, L., Rajamani, R.: A novel wireless piezoelectric tire
  sensor for the estimation of slip angle. Measurement Science and Technology
  21(1),  015201 (2010)

\bibitem{erin2015deep}
Erin~Liong, V., Lu, J., Wang, G., Moulin, P., Zhou, J.: Deep hashing for
  compact binary codes learning. In: Proceedings of the IEEE Conference on
  Computer Vision and Pattern Recognition. pp. 2475--2483 (2015)

\bibitem{Gong13a}
Gong, Y., Kumar, S., Rowley, H.A., Lazebnik, S.: Learning binary codes for
  high-dimensional data using bilinear projections. IEEE Conference on Computer
  Vision and Pattern Recognition pp. 484--491 (2013)

\bibitem{Gong12}
Gong, Y., Kumar, S., Verma, V., Lazebnik, S.: Angular quantization-based binary
  codes for fast similarity search. In: Advances in Neural Information
  Processing Systems. pp. 1196--1204 (2012)

\bibitem{Gong13b}
Gong, Y., Lazebnik, S., Gordo, A., Perronnin, F.: Iterative quantization: A
  procrustean approach to learning binary codes for large-scale image
  retrieval. Pattern Analysis and Machine Intelligence, IEEE Transactions on
  35(12),  2916--2929 (2013)

\bibitem{gong2014multi}
Gong, Y., Wang, L., Guo, R., Lazebnik, S.: Multi-scale orderless pooling of
  deep convolutional activation features. In: Computer Vision--ECCV 2014, pp.
  392--407. Springer (2014)

\bibitem{Gustafsson97}
Gustafsson, F.: Slip-based tire-road friction estimation. Automatica  33(6),
  1087--1099 (1997)

\bibitem{jegou2010aggregating}
J{\'e}gou, H., Douze, M., Schmid, C., P{\'e}rez, P.: Aggregating local
  descriptors into a compact image representation. In: Computer Vision and
  Pattern Recognition (CVPR), 2010 IEEE Conference on. pp. 3304--3311. IEEE
  (2010)

\bibitem{jia2014caffe}
Jia, Y., Shelhamer, E., Donahue, J., Karayev, S., Long, J., Girshick, R.,
  Guadarrama, S., Darrell, T.: Caffe: Convolutional architecture for fast
  feature embedding. In: Proceedings of the ACM International Conference on
  Multimedia. pp. 675--678. ACM (2014)

\bibitem{krizhevsky2012imagenet}
Krizhevsky, A., Sutskever, I., Hinton, G.E.: Imagenet classification with deep
  convolutional neural networks. In: Advances in neural information processing
  systems. pp. 1097--1105 (2012)

\bibitem{Lecun15}
LeCun, Y., Bengio, Y., Hinton, G.: Deep learning. Nature pp. 436--444 (May
  2015)

\bibitem{Leung01}
Leung, T., Malik, J.: Representing and recognizing the visual appearance of
  materials using three-dimensional textons. International Journal of Computer
  Vision  43(1),  29--44 (2001)

\bibitem{Liu_2013_CVPR}
Liu, C., Yang, G., Gu, J.: Learning discriminative illumination and filters for
  raw material classification with optimal projections of bidirectional texture
  functions. In: The IEEE Conference on Computer Vision and Pattern Recognition
  (CVPR) (June 2013)

\bibitem{van2008visualizing}
Van~der Maaten, L., Hinton, G.: Visualizing data using t-sne. Journal of
  Machine Learning Research  9(2579-2605), ~85 (2008)

\bibitem{Matsuzakia05}
Matsuzaki, R., Todoroki, A.: Wireless strain monitoring of tires using
  electrical capacitance changes with an oscillating circuit. Sensors and
  Actuators A: Physical  119(2),  323--331 (2005)

\bibitem{nyahumwa1991friction}
Nyahumwa, C., Jeswiet, J.: A friction sensor for sheet-metal rolling. CIRP
  Annals-Manufacturing Technology  40(1),  231--233 (1991)

\bibitem{perronnin2010large}
Perronnin, F., Liu, Y., S{\'a}nchez, J., Poirier, H.: Large-scale image
  retrieval with compressed fisher vectors. In: Computer Vision and Pattern
  Recognition (CVPR), 2010 IEEE Conference on. pp. 3384--3391. IEEE (2010)

\bibitem{perronnin2010improving}
Perronnin, F., S{\'a}nchez, J., Mensink, T.: Improving the fisher kernel for
  large-scale image classification. In: Computer Vision--ECCV 2010, pp.
  143--156. Springer (2010)

\bibitem{Raginsky09}
Raginsky, M., Lazebnik, S.: Locality-sensitive binary codes from
  shift-invariant kernels. In: Advances in neural information processing
  systems. pp. 1509--1517 (2009)

\bibitem{ramkumar2003developing}
Ramkumar, S., Wood, D., Fox, K., Harlock, S.: Developing a polymeric human
  finger sensor to study the frictional properties of textiles part i:
  artificial finger development. Textile research journal  73(6),  469--473
  (2003)

\bibitem{razavian2014cnn}
Razavian, A.S., Azizpour, H., Sullivan, J., Carlsson, S.: Cnn features
  off-the-shelf: an astounding baseline for recognition. In: Computer Vision
  and Pattern Recognition Workshops (CVPRW), 2014 IEEE Conference on. pp.
  512--519. IEEE (2014)

\bibitem{ILSVRC15}
Russakovsky, O., Deng, J., Su, H., Krause, J., Satheesh, S., Ma, S., Huang, Z.,
  Karpathy, A., Khosla, A., Bernstein, M., Berg, A.C., Fei-Fei, L.: {ImageNet
  Large Scale Visual Recognition Challenge}. International Journal of Computer
  Vision (IJCV) pp. 1--42 (April 2015)

\bibitem{salakhutdinov2009semantic}
Salakhutdinov, R., Hinton, G.: Semantic hashing. International Journal of
  Approximate Reasoning  50(7),  969--978 (2009)

\bibitem{Sharan13}
Sharan, L., Liu, C., Rosenholtz, R., Adelson, E.H.: Recognizing materials using
  perceptually inspired features. International journal of computer vision
  103(3),  348--371 (2013)

\bibitem{shi2012margin}
Shi, Q., Shen, C., Hill, R., Hengel, A.: Is margin preserved after random
  projection? In: Proceedings of the 29th International Conference on Machine
  Learning (ICML-12). pp. 591--598 (2012)

\bibitem{torralba2008small}
Torralba, A., Fergus, R., Weiss, Y.: Small codes and large image databases for
  recognition. In: Computer Vision and Pattern Recognition, 2008. CVPR 2008.
  IEEE Conference on. pp. 1--8. IEEE (2008)

\bibitem{Varma02}
Varma, M., Zisserman, A.: Classifying images of materials: Achieving viewpoint
  and illumination independence. In: Computer Vision—ECCV 2002, pp. 255--271.
  Springer (2002)

\bibitem{vedaldi08vlfeat}
Vedaldi, A., Fulkerson, B.: {VLFeat}: An open and portable library of computer
  vision algorithms. \url{http://www.vlfeat.org/} (2008)

\bibitem{vedaldi15matconvnet}
Vedaldi, A., Lenc, K.: Matconvnet -- convolutional neural networks for matlab
  (2015)

\bibitem{Wang06}
Wang, J., Dana, K.J.: Relief texture from specularities. IEEE Transactions on
  Pattern Analysis and Machine Intelligence  28(3),  446--457 (2006)

\bibitem{Weiss09}
Weiss, Y., Torralba, A., Fergus, R.: Spectral hashing. In: Koller, D.,
  Schuurmans, D., Bengio, Y., Bottou, L. (eds.) Advances in Neural Information
  Processing Systems 21, pp. 1753--1760. Curran Associates, Inc. (2009),
  \url{http://papers.nips.cc/paper/3383-spectral-hashing.pdf}

\bibitem{Yu14}
Yu, F.X., Kumar, S., Gong, Y., Chang, S.F.: Circulant binary embedding. In:
  International Conference on Machine Learning (2014)

\bibitem{Zhang15}
Zhang, H., Dana, K., Nishino, K.: Reflectance hashing for material recognition.
  IEEE Conference on Computer Vision and Pattern Recognition (CVPR) pp.
  3071--3080 (2015)

\bibitem{zhangfast}
Zhang, X., Yu, F.X., Guo, R., Kumar, S., Wang, S., Chang, S.F.: Fast orthogonal
  projection based on kronecker product. International Conference on Computer
  Vision  (2015)

\bibitem{zhou2014learning}
Zhou, B., Lapedriza, A., Xiao, J., Torralba, A., Oliva, A.: Learning deep
  features for scene recognition using places database. In: Advances in Neural
  Information Processing Systems. pp. 487--495 (2014)

\end{thebibliography}
\end{document}